\documentclass{article}

\usepackage{PRIMEarxiv}
\usepackage{CJK} 
\usepackage[utf8]{inputenc} 
\usepackage[T1]{fontenc}    
\usepackage{hyperref}       
\usepackage{url}            
\usepackage{booktabs}       
\usepackage{amsfonts}       
\usepackage{nicefrac}       
\usepackage{microtype}      
\usepackage{fancyhdr}       
\usepackage{graphicx}       
\graphicspath{{media/}}     

\title{
Digestion Algorithm in Hierarchical Symbolic Forests:\\A Fast Text Normalization Algorithm and\\Semantic Parsing Framework for\\Specific Scenarios and Lightweight Deployment
}

\author{
  Kevin You\\
  Project Link: \url{https://blog.csdn.net/m0_62984100/article/details/140054725}
}

\begin{document}
\maketitle

\begin{abstract}
Text Normalization and Semantic Parsing have numerous applications in natural language processing, such as natural language programming, paraphrasing, data augmentation, constructing expert systems, text matching, and more. Despite the prominent achievements of deep learning in Large Language Models (LLMs), the interpretability of neural network architectures is still poor, which affects their credibility and hence limits the deployments of risk-sensitive scenarios. In certain scenario-specific domains with scarce data, rapidly obtaining a large number of supervised learning labels is challenging, and the workload of manually labeling data would be enormous. Catastrophic forgetting in neural networks further leads to low data utilization rates. In situations where swift responses are vital, the density of the model makes local deployment difficult and the response time long, which is not conducive to local applications of these fields. Inspired by the multiplication rule, a principle of combinatorial mathematics, and human thinking patterns, a multilayer framework along with its algorithm, the Digestion Algorithm in Hierarchical Symbolic Forests (DAHSF), is proposed to address these above issues, combining text normalization and semantic parsing workflows. The Chinese Scripting Language "Fire Bunny Intelligent Development Platform V2.0" is an important test and application of the technology discussed in this paper. DAHSF can run locally in scenario-specific domains on little datasets, with model size and memory usage optimized by at least two orders of magnitude, thus improving the execution speed, and possessing a promising optimization outlook.
\end{abstract}


\section{Introduction}
The development of deep learning\cite{hopfield1982neural} has become increasingly prominent with the advancement of computing power and the accumulation of data. While deep learning has achieved remarkable success in natural language processing (NLP), particularly with large language models (LLMs) attempting inference, there are still unacceptable drawbacks in some aspects. According to a paper by Apple Company\cite{mirzadeh2024gsm}, LLMs do not truly reason; they just "memorize" answers and then perform pattern matching. A study published in Scientific Reports\cite{dentella2024testing}, which tested with the prompt like "Steve hugged Molly and Donna was kissed by Molly. In this context, was Molly kissed?" found that the accuracy of large models is random, with significant fluctuations in answers and a lack of strong, consistent responses. This uncertainty and instability have impeded the practical application of deep learning in risk-sensitive areas such as medicine and precision instrument manipulation. Despite the current practicality of AI models, they have not yet reached human-like language capabilities. While these models may outperform humans quantitatively in tests, their answers sometimes exhibit clear non-human errors in language comprehension qualitatively. These indications may suggest a lack of combination operator information for effective grammatical and semantic regulations. Credibility has always been a major issue in deep learning due to its black-box nature\cite{ji2019A}; catastrophic forgetting is also one of the main culprits for low data utilization. Furthermore, deep learning models are dense and require the use of all parts of the models for text normalization every time input data are fed, making it difficult to solve problems satisfying rigid time restrictions, leading to concerns in local deployment and delayed responses. Additionally, quickly obtaining a large amount of data for certain Scenario-Specific domains can be challenging, leading to the demand for manual data labeling. Nevertheless, deep learning is a big data method, making manual data labeling a tremendously laborious chore. In lieu of tediously labeling a large amount of data manually, it is more effective to directly tackle these problems by setting rules, which increases universality and achieves more with less effort.
\section{Research Subject}
The research focuses on text normalization and semantic parsing, which have numerous uses in natural language processing. One of the most notable applications is their ability to bridge between advanced programming language and natural language\cite{chen2021evaluating} by matching synonymous sentences. Apart from this, there are various other wide-ranging employments. For example, by combining different language metrics (emotions, language styles, etc.), text normalization enables computers to produce statements in given styles or to modify wording. It can also be used for data augmentation to generate training sets, even when data is limited, by extending a single sentence to all similar sentences, thus enhancing data diversity. Text normalization is also a technology that boosts machine thinking\cite{ye2023large} by assessing the extent, to which external conditions match existing knowledge or not, and deducing conclusions from existing knowledge bases with the assistance of reasoning engines like Production Systems\cite{newell1963gps}. Furthermore, this technique may aid in text matching\cite{jing2017building}, for instance in search engines and recommendation algorithms.

This article aims to construct a lightweight deployable text normalization and semantic parsing algorithm called the "Digestion Algorithm in Hierarchical Symbolic Forests (DAHSF)", a fast algorithm for connecting free sequences to highly standardized sequences. The algorithm takes a finite-length sequence as input (e.g., "\begin{CJK*}{UTF8}{gkai}鼠标左键双击（19，16），随后等待7秒，复制当前选中内容，缩小页面，关闭当前页面，并按下回车键\end{CJK*}") and outputs a finite-length regularized sequence (e.g., "\begin{CJK*}{UTF8}{gkai}双击（19，16），等7秒，拷，缩小，关页，回车\end{CJK*}"), which is an equivalent expression of the input but in a standardized and relatively deterministic form. The precursor project has simplified its usage as much as possible, making it intuitive and clear. Therefore, this article can present an algorithm for text normalization (i.e., a synonym sentence transformation algorithm under the same semantics) and semantic parsing (including tokenization) algorithms.
\begin{figure}[h]
  \centering
  \includegraphics[width=\linewidth, trim=0mm 140mm 0mm 0mm, clip]{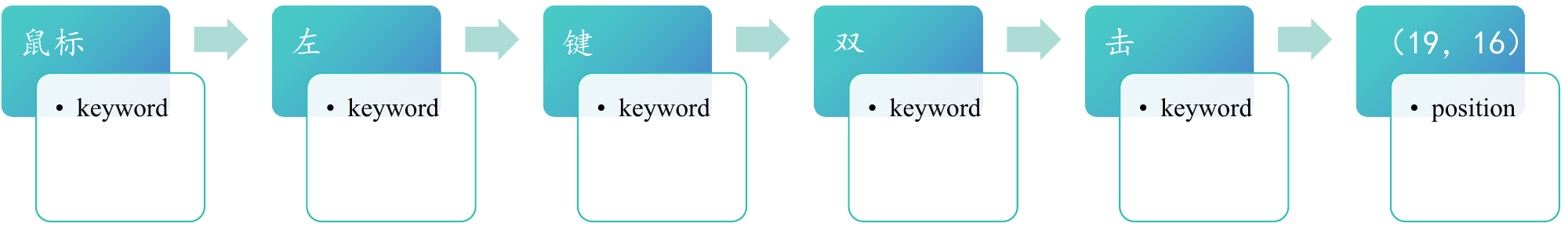}
  \caption{Performing semantic parsing when tokenizing.}
  \label{fig:fig1}
\end{figure}
\section{Problem Formulation}
To prevent ambiguity, let us redefine some important concepts in this article and make necessary assumptions.

There are multiple dimensions to analyze the semantics of texts, such as actions, emotions, and the formality of expression. We will use the term "semantic dimensions" to represent them. For example, when connecting the advanced Chinese scripting language in the predecessor project "Fire Bunny  Intelligent Development Platform V1.0" to natural language, controlling the semantic dimension "action information" is crucial for synonymous transformations. Focused on the speaker's requests and commands, if two text segments convey the same action information, regardless of other irrelevant details like the speaker's tones and emotions, they are said to be "action-synonymous". Likewise, we can define emotion-synonymous and more. Breaking down text semantics into multiple dimensions facilitates language transformations in specific styles.

A sentence may be transformed into multiple predetermined sequence segments. If a sentence can only be transformed into one specific predetermined segment, it is defined as a "simple sentence". For example, in our precursor project, "\begin{CJK*}{UTF8}{gkai}最大化\end{CJK*}" can only be transformed into "\begin{CJK*}{UTF8}{gkai}最大化\end{CJK*}". Hence, "\begin{CJK*}{UTF8}{gkai}最大化\end{CJK*}" is a simple sentence. Therefore, in terms of semantic equivalence, all compound sentences are equivalent to a graph with corresponding simple sentences as nodes and logical structures as edges. Synonymous simple sentences can be transformed into each other through 1) synonym replacement and 2) grammatical structure transformations maintaining conceptual identity and sentence structural parallelism.

We further abstract the elements of sequences. From the following context, "Word" and "Sentence" are defined as follows.
\paragraph{Definition 3.1: Word}
A \textit{word} is a sequence segment with a specific meaning that originates from a division of a sequence according to certain rules.

An $n$-length sequence $e_1e_2\cdots e_n$ is divided into $(e_1,\cdots,e_{i_1}),\cdots,(e_k,\cdots,e_{i_k})$ according to predefined principles, where $i_k=n$. Then $(e_j,\cdots,e_{i_j}),~1\le j\le k$ is a \textit{word}.

Take a sequence of tuples as an example:

\begin{center}
    $(1, 2, 3)(4, 5, 6)(2, 3, 4)(5, 6, 7)$
\end{center}

If we join the tuples together to form \textit{words} according to the rule of increasing numbers between consecutive tuple intervals (in this case: 3<4 and 4<5), then both "$(1, 2, 3)(4, 5, 6)$" and "$(2, 3, 4)(5, 6, 7)$" are \textit{words} of the above sequence.
\paragraph{Definition 3.2: Sentence}
A \textit{sentence} is a specific \textit{word} sequence that adheres to certain rules.

If $S=(e_{j_1},\cdots,e_{i_{j_1}}),\cdots,(e_{j_m},\cdots,e_{i_{j_m}})$ obeying preset principles, then $S$ is a \textit{sentence}.

For instance, if we define that a \textit{sentence} is exactly composed of 4 \textit{words}, then
\begin{center}
    $(1, 2, 3)(4, 5, 6)$\\
    $(2, 3, 4)(5, 6, 7)$\\
    $(2, 3, 4)(5, 6, 7)$\\
    $(1, 2, 3)(4, 5, 6)$\\
\end{center}
is a \textit{sentence}. We denote it as $S_0$. The \textit{sentences} at the original level can be reconsidered as \textit{words} when we elevate our horizons to a higher level. Upon ascending a level, the \textit{sentences} form a sequence, such as the repetition of $S_0$ five times:
\begin{center}
    $C:=S_0S_0S_0S_0S_0$
\end{center}
Now, we can consider the original \textit{sentence} of the original level $S_0$ as a \textit{word} at a new higher level.

When discussing \textit{words} and \textit{sentences}, it is advisable to indicate the level they belong to, unless the context makes it explicitly clear.

This article aims to transform any sequence into a highly standardized form, i.e., a sequence of preset contents, or to identify a \textit{word} beyond knowledge. Now we introduce two important concepts of highly normalized sequences.

\paragraph{Definition 3.3: Sentence Framework}
A \textit{sentence framework} is a fixed pattern with predetermined vacancies that, after filling these vacancies with specified types of \textit{words}, a \textit{sentence} is composed.
\paragraph{Definition 3.4: Data Word}
A \textit{data word} is a \textit{word} whose patterns and contents may be diverse but can be classified by identifiers, though it can be hard to predict directly without given contexts.
\paragraph{Definition 3.5: Keyword}
\textit{Keywords} are the \textit{words} but not vacancies that compose \textit{sentence frameworks}.

Given the abundance of \textit{data words}, it is often necessary to categorize them. To classify them, we shall label them and indicate their types (they are the \textit{subclasses} of \textit{data words}).

In Python codes, for instance, the following commands can be found:
\begin{center}
import numpy as np
\end{center}
From this, we can extract the \textit{sentence framework}:
\begin{center}
import [module name] as [alias]
\end{center}
In this text, "import" and "as" are \textit{keywords}, while "numpy" and "np" represent the module name and the alias of the module, respectively. Both "numpy" and "np" are \textit{data words}. Similarly, we can have:
\begin{center}
import tensorflow as tf\\
import matplotlib.pyplot as plt
\end{center}
Both of their \textit{sentence frameworks} are:
\begin{center}
import [module name] as [alias]
\end{center}
\textit{Data words} (i.e, [module name][alias]) respectively, are:
\begin{center}
"tensorflow" "tf" and "matplotlib.pyplot" "plt"\\
\end{center}
Again, they share the same \textit{keywords} "import" and "as".

Given the focus on text normalization and semantic parsing in scenario-specific domains, it is possible to establish scenario-specific lexicons tailored to the target field. For example, to include all Python built-in keywords and local third-party library nouns. The effort required to construct lexicons in this manner, compared to supervised learning methods that involve manual data labeling, is sometimes not enormous.
\section{Digestion Algorithm in Hierarchical Symbolic Forests}
This article aims to perform text normalization, which outputs an equivalent form of the input, and semantic parsing. Under the transformation of the Digestion Algorithm in Hierarchical Symbolic Forests, each word of this output sequence either represents a keyword, a data word, or an unknown word beyond knowledge.

Since our task is to convert versatile sequences into standardized ones and, inspired by the multiplication rule in combinatorial mathematics, we think that it requires a hierarchical architecture to solve this problem. Diverse sentence forms can be generated step by step through the combination of different levels. Similar to a neural network, our architecture can be represented as a graph with nodes and edges. However, the lack of interpretability in neural networks stems from the inexplicability of their nodes and edges.

Drawing inspiration from human thinking patterns\cite{mcculloch1943logical}\cite{wu2023graph} and artificial neural networks\cite{rumelhart1986learning}, in each layer of our architecture, we assign meanings to all edges and nodes that are relevant to the input data of that layer. By linking an edge with a mapping and binding every object (or word) to each node, we aim to eliminate interpretability issues fundamentally.

Synonym matching is a process of partitioning into equivalence classes at the word level, where multiple different words often belong to the same equivalence class. Motivated by this fact, this article introduces the architecture of Hierarchical Symbolic Forests (HSF). The process of "digestion" is the process in which free texts are firstly "swallowed" and then gradually transformed into increasingly digestible forms within different "organs". Through Hierarchical Symbolic Forests, we achieve synonym matching, and through the Digestion Algorithm (DA), we accomplish semantic parsing.

To match synonyms, it is often necessary to know the type of a word - either it is a keyword or a data word. Given a data word, which subclass it belongs to? For this purpose, we have innovated a tokenization method based on a lexicon. By combining Hierarchical Symbolic Forests with the Digestion Algorithm, we parse semantics layer by layer, capturing the structural information of language from lower to higher levels, and progressively transforming free text into more standardized sequences. This approach allows us to peel back the layers of hierarchical information in reality.

\begin{figure}[h]
  \centering
  \includegraphics[width=\linewidth, trim=0mm 158mm 0mm 0mm, clip]{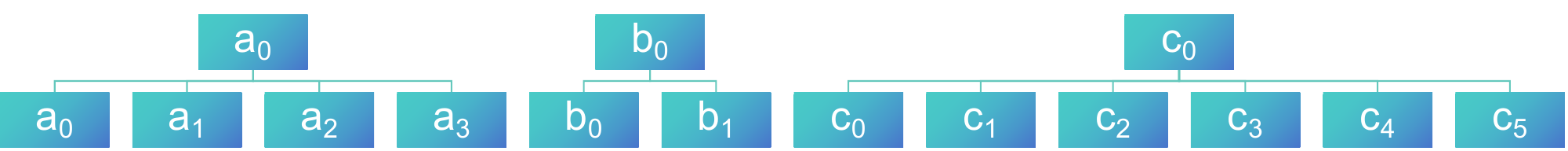}
  \caption{The graph illustrates the structure of one layer of Hierarchical Symbolic Forests. Here, $a_0$, $b_0$, and $c_0$ serve as representative elements, each is represented as a root node, with a lower level consisting of all elements within their respective equivalence classes, depicted as leaf nodes.}
  \label{fig:fig2}
\end{figure}

Hierarchical Symbolic Forests are layers of forests, as shown in Figure \ref{fig:fig2}, where each node is endowed with a meaning, hence the name "symbolic". For instance, if the input sequence for the first layer is "$a_1$, $c_5$", then the first layer would output sequence "$a_0$, $c_0$", where "$a_0$" is equivalent to "$a_1$" and "$c_0$" is equivalent to "$c_5$", achieving the first tokenization step as well. Subsequently, the types of words are identified and information is injected into the output of the first layer: this is the process of digestion. The Digestion Algorithm represents the cumulative effect of this "digestion" at each layer. Words form sentences, and when viewed from a higher level, sentences are words of a new level. Therefore, this architecture can be chained multiple times, each handling sequences at different levels. This is the essence of Hierarchical Symbolic Forests.

Now we deploy the Digestion Algorithm to Hierarchical Symbolic Forests. Concretely, let the $i$th layer $(i\ge1)$ of Hierarchical Symbolic Forests be the mapping $L_i$, which relies on the equivalence relation $R_i$ from the lexicon of this layer $S_i$ (notably, this set can be countable but infinite, as data words could be infinite), mapping the word $x_i$ to its representative element $\overline{x_i}$. Then, $\overline{x_i}$ is assigned with a certain meaning called label $l_{(i,\overline{x_i})}$ (or an empty label). The process of labeling and packing a word into a tuple is the "digestion" process, which constitutes the Digestion Algorithm. Denote $[\overline{x_i}]$ as the set of all words equivalent to $x_i$, satisfying $[\overline{x_i}]\subset S_i$. We also have $l_{(i,\overline{x_i})}\in l_i$, in which $l_i$ is the label set of this layer. Combining all of these, it yields:
\begin{equation}
L_i(R_i):x_i\to(\overline{x_i}, l_{(i,\overline{x_i})}),\forall x_i\in S_i
\end{equation}
Let $x_{i+1}=[(\overline{x_{i_1}},l_{(i_1,\overline{x_{i_1}})}),\cdots,(\overline{x_{i_j}},l_{(i_j,\overline{x_{i_j}})})](j\ge1)$, a sentence of layer $i$ and meanwhile, a word of layer $i+1$. Then we can construct $L_{i+1}$,$L_{i+2}$, etc.

With the example of the "Fire Bunny Intelligent Development Platform V2.0", the method to build Hierarchical Symbolic Forests architecture and to deploy the Digestion Algorithm will be illustrated below. Two layers are used in the input pipeline of this platform.

Set separately, the first layer prioritizes longer (literally) words [longer segments often carry unique meanings; e.g., "\begin{CJK*}{UTF8}{gkai}火箭\end{CJK*}" (rocket) and "\begin{CJK*}{UTF8}{gkai}火\end{CJK*}" (fire) are two distinct entities, not a simple addition of "\begin{CJK*}{UTF8}{gkai}火\end{CJK*}" and "\begin{CJK*}{UTF8}{gkai}箭\end{CJK*}" (arrow)] in respective lexicon. At the first layer, this platform utilizes its lexicon to tokenize texts for word normalization of the initial phase. This step functions as a "perceiver", first receiving contents based on the lexicon. Initially, thesaurus forests\cite{tian2010words} $L_1$ are established using Hierarchical Symbolic Forests, turning diverse synonyms $x_1$ into uniform core synonyms $\overline{x_1}$, the shortest word in $S_1$. $L_1$ matches $\overline{x_1}$ to itself when inputting the core synonym. In this phase, the platform will digest what can be primarily digested (i.e., finding the keywords that compose a sentence framework). After this digestion, it extracts data words and determines their subclasses from the remaining indigestible portion. This process initially segments sentences and promptly labels these pieces.

This implies a methodology. By establishing rules with prepared knowledge (setting $L_1$), sentences are digested into individual words and reassembled into a more orderly standardized form. Perhaps elevating to higher dimensions without modifying the original string is one of the best ways to label strings. In this platform, after processing, "www.baidu.com" segment becomes (\begin{CJK*}{UTF8}{gkai}“\end{CJK*}www.baidu.com\begin{CJK*}{UTF8}{gkai}”\end{CJK*}, \begin{CJK*}{UTF8}{gkai}网址\end{CJK*}), where "\begin{CJK*}{UTF8}{gkai}网址\end{CJK*}" stands for URLs in Chinese, and, as a preset rule in the "Fire Bunny Intelligent Development Platform V1.0", Chinese quotation marks must be added around "www.baidu.com". In the next layer, (\begin{CJK*}{UTF8}{gkai}“\end{CJK*}www.baidu.com\begin{CJK*}{UTF8}{gkai}”\end{CJK*}, \begin{CJK*}{UTF8}{gkai}网址\end{CJK*}) constitutes the smallest unit of any new sequence. Units like these compose a new sequence. Tokenization is a specialized case of this approach. Correctly segmenting a sentence (especially in languages like Chinese without spaces between words) involves initial digestion and later processing. The tokenization process is also the labeling (parsing) process - simply adding a dimension to the processed segments.

At layer 2, the sequence outputted from the first layer undergoes a secondary transformation. Subjectively, we introduce this higher-level transformation to allow advanced usages in Chinese like omission and conversion. During this process, ambiguity is initially removed (parsing commands), followed by contextual analysis (second synonymous matching).

\begin{figure}[h]
  \centering
  \includegraphics[width=\linewidth, trim=0mm 132.5mm 197mm 0mm, clip]{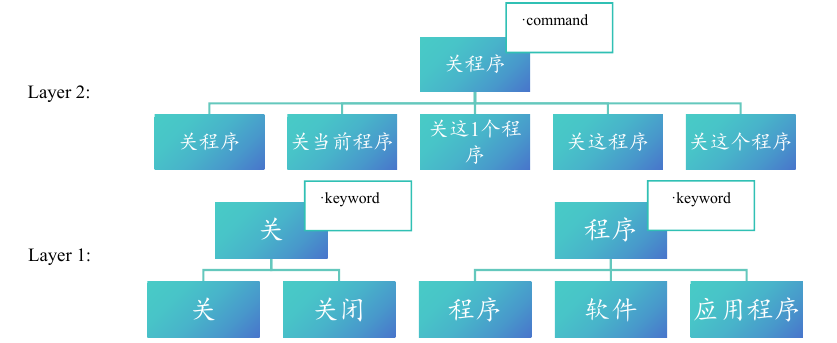}
  \caption{The HSF involved in the "\begin{CJK*}{UTF8}{gkai}关程序\end{CJK*}" command.}
  \label{fig:fig3}
\end{figure}

For instance, the command "\begin{CJK*}{UTF8}{gkai}关程序\end{CJK*}" in the "Fire Bunny Intelligent Development Platform V2.0" has various expressions. See Figure \ref{fig:fig3}. There are at least $5\times2\times3=30$ ways to express this command. For illustrative purposes, we omit the words that are matched to themselves in thesaurus forests at layer 1 (keywords "\begin{CJK*}{UTF8}{gkai}当前\end{CJK*}", "\begin{CJK*}{UTF8}{gkai}这\end{CJK*}", "\begin{CJK*}{UTF8}{gkai}个\end{CJK*}", and "1" can only be matched themselves). In layer 1, every appearing word (e.g., "\begin{CJK*}{UTF8}{gkai}关\end{CJK*}", "\begin{CJK*}{UTF8}{gkai}程序\end{CJK*}", including the ones aforementioned) is a keyword; thus, we will no longer label with "keyword" within the leaf nodes in layer 2.

\section{Evaluations and Further Analyses}
By establishing "Fire Bunny Intelligent Development Platform V2.0", we have deployed the DAHSF to a local laptop. To obtain a test set, we first simulate the correct answers and then use ChatGLM-4 to simulate user inputs in batches.

Dealing with long corpus, our local lightweight architecture outperforms both cloud and locally deployed large language models. Specifically:

\begin{itemize}
\item Text normalization and semantic parsing are rapid to the point of being nearly imperceptible. Whereas, the runtime of locally deployed large language models is a disaster. Even in the cloud, large language models have a noticeable time lag.
\item There is almost no limit on the input length, which distinguishes our framework significantly from large language models.
\end{itemize}
The model is tiny, taking up approximately 1,589,763 bytes when stored on disks, and memory usage does not exceed 10MB when parsing small texts. In contrast, a GPT-based model may require memory in gigabyte magnitude\footnote{The data in Table \ref{tab:table}, except HSF, are all generated by the ChatGLM-4.}, and will not achieve good accuracy unless the model is sufficiently large. See Table \ref{tab:table} for more details.

\begin{table}
 \caption{A comparison between DAHSF and existing large language models}
  \centering
  \begin{tabular}{ccccc}
    \toprule
    Models    & HSF    & Gemini    & GLM4    & GPT4-mini\\
    \midrule
    Is Execution Speed Perceptible?    & No    & Yes    & Yes    & Yes\\
    Order of Magnitude of Model Size    & 1 MB    & 1 GB    & 1 GB    & 100 MB\\
    Order of Magnitude of Memory Consumption    & 10 MB    & 10 GB    & 10 GB    & 1 GB\\
    Optimized Model's Part    & Structure    & Parameters    & Parameters    & Parameters\\
    \bottomrule
  \end{tabular}
  \label{tab:table}
\end{table}
Meanwhile, we have updated our "Fire Bunny Intelligent Development Platform V1.0", which has obtained the "Computer Software Copyright Registration Certificate" from the National Copyright Administration of the People's Republic of China, with certificate number 13272886, introducing the following new features:

\begin{center}
Remote control, custom commands, interaction with other programming languages, numerical calculations, screenshot abilities, simple local file management, chat history, streaming answers, software interface, and natural language programming.
\end{center}

The DAHSF has been locally deployed and successfully integrated as an input pipeline in the "Fire Bunny Intelligent Development Platform V2.0", further bringing natural language programming closer to realization.

For example, to open the Baidu homepage, use the code "\begin{CJK*}{UTF8}{gkai}打开www.baidu.com\end{CJK*}". Through the first layer of HSF, the algorithm first detects the keyword "\begin{CJK*}{UTF8}{gkai}打开\end{CJK*}" (open), separates it from www.baidu.com, then converts it into a more concise expression "\begin{CJK*}{UTF8}{gkai}开\end{CJK*}" (open), and identifies the subclass of data word "www.baidu.com" as a URL. With the sentence framework matching in the second layer, "\begin{CJK*}{UTF8}{gkai}打开【网址】\end{CJK*}" is transformed into "\begin{CJK*}{UTF8}{gkai}开【网址】\end{CJK*}". Combined with the data word "www.baidu.com", the command "\begin{CJK*}{UTF8}{gkai}开www.baidu.com\end{CJK*}" is generated, instructing the computer to open the Baidu official website. In another case on laptops with Windows 11 Operation Systems, to open WPS software (if already installed), the code can be "\begin{CJK*}{UTF8}{gkai}启动WPS\end{CJK*}". Here, "\begin{CJK*}{UTF8}{gkai}启动\end{CJK*}" (start) is also a keyword, which is matched with "\begin{CJK*}{UTF8}{gkai}开\end{CJK*}" (open). "WPS" is a known program name, a subclass of data words. The sentence framework "\begin{CJK*}{UTF8}{gkai}开【程序名】\end{CJK*}" in the second layer is activated, prompting the computer to execute the operation of "\begin{CJK*}{UTF8}{gkai}启动WPS\end{CJK*}".
\section{Prospects}
Based on the application purposes, we will focus on two main directions: natural language programming\cite{liu2024autoglm} and the application of production systems with a self-learning mechanism.

When NLP empowers advanced programming languages like in the "Fire Bunny Intelligent Development Platform V1.0", programming paradigms can transcend from high-level language programming to the realm of natural language programming, using a single line of Chinese command can control mouse and keyboard operations, open specific applications with a simple command, or manipulate software interfaces and systems with ease. Such exploration is conducive to discovering more concise and powerful human-computer interaction methods, providing more references and support for building other human-computer interfaces and other artificial intelligence technologies. The natural language programming paradigm is poised to challenge existing programming patterns. On the one hand, with text normalization and semantic parsing pipelines, automating office tasks will become seamless - write down the workflows and execute them directly, significantly improving work efficiency. Natural language programming facilitates automated office tasks, enabling humans to control computers deeply, laying a crucial foundation for deep manipulation of the external world, and freeing humans from repetitive, tedious, and potentially life-threatening tasks. On the other hand, breaking the monopoly of English computer languages will lower the barriers to programming. For people unfamiliar with English, the ability to control computers deeply will be a piece of cake, and sharpening computer skills will no longer depend on proficiency in English. Particularly for younger students, natural language programming is an effective way to quickly cultivate programming, interdisciplinary, and AI thinking, with significant educational value. In the future, as AI impacts the world more directly (there would be more avenues of influence with the assistance of multimodal models\cite{niu2024screenagent}\cite{lu2024omniparser}), compared to traditional chatbots that must influence human thinking before impacting the world, AI will serve as a new productive force to elevate living standards and stimulate a new round of economic growth.

The second purpose of the design in this article is to understand human thinking and decision-making patterns, to comprehend why computers think the way they do, to enhance the interpretability of computer thinking, to allow humans to understand computer decisions when it comes to teaching students computers, to identify areas where computers may make thinking mistakes, etc. By correcting computer-made errors, the credibility of computer decisions will be increased, achieving mutual enhancement and efficient collaboration between humans and computers. Knowledge-based decision-making systems hold the potential to revolutionize the learning paradigm of deep learning and thoroughly address the challenges of interpretability, the problem of catastrophic forgetting, and long local inference time issues in deep learning. In traditional neural networks, there is a lack of clear binary opposition between connections, which, to some extent, may contribute to the looseness of neural networks' inference. The interpretability issue of neural networks stems from the interpretability issues of individual neurons and edges (connections between neurons). If each neuron corresponds to a specific condition (encoding neurons as switches that determine whether a condition is met), such as "[One object] is a particle", and connects the neurons through logical rules (e.g., "and", "or", "not"), the black box problem of neural networks may be fundamentally addressed. In conjunction with inference engines and large models, lighting up a "knowledge" (i.e., satisfying condition of certain theorems in the theorem library) requires mapping natural language to formatted text "conditions". This process relies on text normalization, a technology that maps unstructured external observations to our knowledge bases. Knowledge-based decision-making systems (such as expert systems) essentially consist of two: condition matching and inference engines. The architecture outlined in this article helps to handle the condition-matching problem. Additionally, the lexicon can be the foundation for making decisions and act as a "memory repository". Thus, for machines, effective "learning" should involve adding new content to the lexicon and correcting potential errors.

\section{Future Works}
Every framework has imperfections: without exception, there is room to optimize this architecture as well. This framework currently lacks a "learning mechanism" to improve its performance, with the model's performance being entirely dependent on its structural design. So far, the knowledge base has not been able to update automatically, thus the model's performance. Therefore, in the upcoming "Fire Bunny Intelligent Development Platform V3.0", we will explore automated learning algorithms. For instance, we may utilize generative AI to create test cases and consider testing "hypothesis" to automatically edit the knowledge base, thereby designing algorithms to train models of this type. Furthermore, the errors made by the model in test cases exhibit significant commonalities. In the future, we will automate the process of model error analysis, knowledge base updates, and building logic frameworks to lay the foundation for the model's self-learning and self-optimization, while maintaining a high level of interpretability.

Currently, we have not considered situations where users input incorrectly. We aim to solve this issue in the third edition and design a more detailed classification and parsing framework. In the "Fire Bunny Intelligent Development Platform V3.0", we will establish a mechanism for autonomous learning. To enable themselves to truly "extract knowledge", machines should "practice" in various real-world scenarios, use generative techniques to generate "assumptions", test their validity, retain those that perform well in the dataset as knowledge, and question existing hypotheses that may be incorrect. The theorem base works similarly. So does the lexicon. Only through this approach can we create an AI brain that "learns on its own", capable of sharing knowledge with us.

\end{document}